\documentclass[journal]{IEEEtran}
\usepackage{amsmath,amsfonts}
\usepackage{algorithmic}
\usepackage{algorithm}
\usepackage{array}
\usepackage[caption=false,font=normalsize,labelfont=sf,textfont=sf]{subfig}
\usepackage{textcomp}
\usepackage{stfloats}
\usepackage{url}
\usepackage{verbatim}
\usepackage{graphicx}
\usepackage{subcaption}
\usepackage{multirow}

\usepackage{cite}
\begin{document}

\title{AD-DINO: Attention-Dynamic DINO for Distance-Aware Embodied Reference Understanding}

\author{Hao Guo, Wei Fan, Baichun Wei, ~\IEEEmembership{Member,~IEEE}, Jianfei Zhu, Jin Tian,\\ Chunzhi Yi, ~\IEEEmembership{Member,~IEEE}, Feng Jiang, ~\IEEEmembership{Senior Member,~IEEE}
        % <-this % stops a space
\thanks{This work is founded by National Natural Science Foundation of China (No.62076080, 62306083), Natural Science Foundation of ChongQing CSTB2022NSCQ-MSX0922 and the Postdoctoral Science Foundation of Heilongjiang Province of China (LBH-Z22175).(Corresponding author: Chunzhi Yi, Feng Jiang.)}% <-this % stops a space
\thanks{Hao Guo and Jianfei Zhu are with the Faculty of Computing, Harbin Institute of Technology, Harbin 150001, China (e-mail: guohao@hit.edu.cn, nyzhujianfei@stu.hit.edu.cn).}
\thanks{Wei Fan and Jin Tian are with the School of Mechatronics Engineering, Harbin Institute of Technology, Harbin 150001, China (e-mail: 23S136392@stu.hit.edu.cn, jin.tian@stu.hit.edu.cn).}
\thanks{Baichun Wei and Chunzhi Yi are with the School of Medicine and
Health, Harbin Institute of Technology, Harbin 150001, China (e-mail:
bcwei@hit.edu.cn, chunzhiyi@hit.edu.cn).}
\thanks{Feng Jiang is with the Faculty of Computing, Harbin Institute of
Technology, Harbin 150001, China, and with the School of Artificial Intelligence, Nanjing University of Information Science \& Technology, Nanjing 210044, China (e-mail: fjiang@hit.edu.cn).}
}

\markboth{Journal of \LaTeX\ Class Files,~Vol.~14, No.~8, October~2024}%
{Shell \MakeLowercase{\textit{et al.}}: A Sample Article Using IEEEtran.cls for IEEE Journals}


\maketitle

\begin{abstract}
Embodied reference understanding is crucial for intelligent agents to predict referents based on human intention through gesture signals and language descriptions. This paper introduces the Attention-Dynamic DINO, a novel framework designed to mitigate misinterpretations of pointing gestures across various interaction contexts. Our approach integrates visual and textual features to simultaneously predict the target object's bounding box and the attention source in pointing gestures. Leveraging the distance-aware nature of nonverbal communication in visual perspective taking, we extend the virtual touch line mechanism and propose an attention-dynamic touch line to represent referring gesture based on interactive distances. The combination of this distance-aware approach and independent prediction of the attention source, enhances the alignment between objects and the gesture represented line. Extensive experiments on the YouRefIt dataset demonstrate the efficacy of our gesture information understanding method in significantly improving task performance. Our model achieves 76.4\% accuracy at the 0.25 IoU threshold and, notably, surpasses human performance at the 0.75 IoU threshold, marking a first in this domain. Comparative experiments with distance-unaware understanding methods from previous research further validate the superiority of the Attention-Dynamic Touch Line across diverse contexts.
\end{abstract}

\begin{IEEEkeywords}
Embodied reference understanding, visual grounding, referring expression comprehension.
\end{IEEEkeywords}

\section{Introduction}
\IEEEPARstart{R}{eference} understanding (RU) is fundamental to interpersonal and human-robot communication, particularly when referring to shared objects within a common space \cite{stacy2020intuitive, tang2020bootstrapping}. In the computer vision community, referring expression comprehension (REC) \cite{ye2019cross,liu2019improving,yu2018mattnet,qiu2020human,fan2021learning, lu2024lgr,shang2022cross}, a crucial task within RU, facilitates visual grounding in images by integrating visual and linguistic cues. However, accurately locating referents based solely on verbal descriptions remains challenging \cite{qiao2020referring,liu2019learning,hu2017modeling,qiu2024mcce,li2023fully}. Non-verbal references, such as embodied pointing gestures synchronized by the describer, often provide more precise spatial indication than verbal descriptions alone. To enhance the combined interpretation of linguistic and visual cues, Chen et al.\cite{chen2021yourefit} introduced the embodied reference understanding (ERU) task, along with the YouRefIt dataset and benchmark. Nevertheless, the interpretation of pointing gestures continues to significantly constrain their effectiveness in ERU applications.

The comprehension of pointing gestures is deeply rooted in human cognitive development and learning processes \cite{liszkowski2004twelve,liszkowski200612}. Traditionally, it has been assumed that the target object's orientation lies within the extension of the pointer's arm-finger line. Intriguingly, recent studies have revealed that this arm-finger line represents a systematic spatial misunderstanding \cite{herbort2016spatial,herbort2018point}. When the referent is distant from the pointer, observers can utilize the eye-finger line — formed by aligning the eye, finger, and referent — to mitigate misinterpretation of the pointing gesture. Building on this concept, O'Madagain et al. \cite{o2019origin} proposed the Virtual Touch Line (VTL) mechanism, which Li et al.\cite{li2023understanding} subsequently employed to connect the eye to the fingertip, significantly enhancing ERU performance.

\begin{figure}
\centering
\includegraphics[width=3.4in]{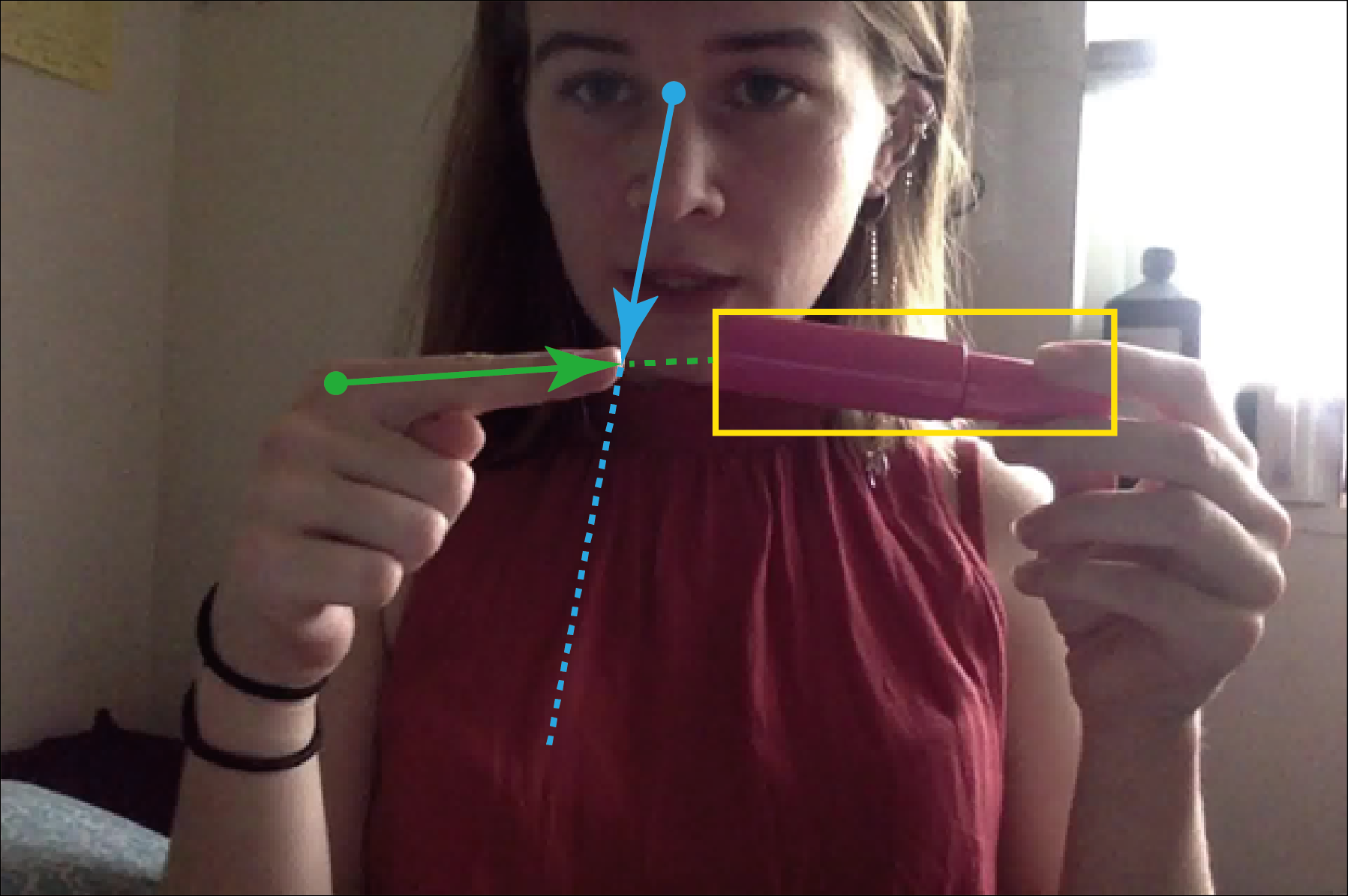}
\caption{Example of pointing to the object that close to body. In this scenario, the girl is accurately pointing at the lipstick which is around her while the eyes deviate the connect line between fingertip and lipstick.}
\label{fig_1}
\end{figure}

However, as pointing is a situated interactive activity \cite{kita2003pointing}, the VTL mechanism can be misinterpreted in close-proximity interactions. Fig. \ref{fig_1} illustrates this limitation: a girl points to a nearby lipstick, yet the VTL and its extension do not intersect with the object. In this scenario, previously used reference lines such as the wrist-elbow or arm-finger lines are not clearly identifiable. The pointer's limb may bend arbitrarily while maintaining an extended index finger, allowing the finger line (FL) to pass through the object's spatial location, thus providing a clear embodied reference.

Embodied reference understanding is predicated on visual perspective taking (VPT)\cite{qiu2020human,batson1997perspective,michelon2006two}, where mutual engagement requires both parties to adjust their cognitive and perceptual states for alignment. We argue that previous research has focused mechanistically on interpreting gesture appearances, neglecting the pointer-referent interaction contexts that inform gesture formation from the pointer to the observer. Specifically, pointers invariably aim to provide clear gestures by directing their finger toward the target object. In distant interactions, pointers spatially align their eye, finger, and the object to accurately indicate the referent's position, adhering to the VTL mechanism. However, in close interactions, pointers disregard VTL-imposed alignment restrictions, allowing for more casual yet accurate pointing under gaze supervision. While O'Madagain et al.'s theory of pointing originating from touch \cite{o2019origin} remains valid, the VTL mechanism becomes inapplicable in these scenarios. Drawing inspiration from multiple studies \cite{scott1997reaching, sabes2000planning, zaepffel2012planning} and common sense, which noted that humans adjust arm positions during pointing movements through complex motion planning based on object position and distance, we introduce the concept of distance-aware VPT (DA-VPT). This approach extends gesture representation from VTL to the Attention-Dynamic Touch Line (ADTL). In the ADTL framework, while the endpoint remains fixed at the fingertip, the attention source (starting point) becomes dynamic, adapting to the interaction distance. For distant interactions, the attention source is set to the eyes, aligning with the VTL. However, in close-proximity interactions, considering DA-VPT, we shift the attention source to the metacarpophalangeal joint (MCP) of the index finger, corresponding to the FL, resulting in a clearer gesture representation compared to the VTL.

To address the limitations of previous research and incorporate the ADTL concept, we propose the Attention-Dynamic DINO (AD-DINO) to enhance visual grounding accuracy. Our system processes visual and natural language inputs through initial feature extraction, cross-modal feature fusion, and language-guided query selection for the decoder. Both language and visual features are simultaneously fed into the cross-modality decoder, which directly outputs the object's bounding box location and the attention source. Additionally, we integrate a fingertip detector to determine the fingertip's position, enabling the construction of the ADTL corresponding to the human body pose by combining the attention source and fingertip position. The selection of the attention source is based on the interaction distance between the pointer and the object. When this distance exceeds arm's length, we localize the attention source to the eyes. Otherwise, we shift the attention source from the eyes to the MCP. Independent prediction of the attention source also reduces model training costs and error levels compared to attention source -fingertip pair prediction.

By the designed AD-DINO and ADTL, our proposed method achieves 76.3\% accuracy under the 0.25 IoU threshold. Significantly, at the 0.75 IoU threshold, AD-DINO achieves 55.4\% accuracy and outperforms human performance for the first time in embodied reference understanding tasks, marking a crucial milestone in narrowing the gap between computational models and human capabilities in this domain.

The main contributions of this paper are fourfold:
(i) We emphasize the distance-aware aspect of the visual perspective taking mechanism and incorporate it into the embodied reference understanding task.
(ii) We develop an attention-dynamic touch line based on pointer-referent interaction distance and distance-aware visual perspective taking, enhancing embodied reference understanding performance.
(iii) We propose a novel model that combines images with gesture information and verbal instructions, optimizing the detection mechanism for key points in pointing gestures.
(iv) We achieve state-of-the-art (SOTA) performance on ERU, with our method demonstrating a 16.4\% improvement over the previous SOTA method at the 0.75 IoU threshold on the YouRefIt dataset, surpassing human performance for the first time.

\section{RELATED WORK}
\subsection{Interpretation of Pointing Gesture}
Pointing gestures serve as a nonverbal communication method to indicate directions, objects, or attract attention. Traditionally, these gestures were interpreted from the observer's perspective using the extension of the pointer's arm-finger line\cite{lucking2015pointing}. However, recent studies have characterized this approach as an empirical and systematic spatial misunderstanding \cite{herbort2016spatial,herbort2018point}. Recent research suggests that pointers indicate objects using an eye-finger-object aligned Virtual Touch Line. Herbort \& Kunde \cite{herbort2018point} demonstrated that employing the VTL extension enhances referent localization accuracy. O'Madagain et al. \cite{o2019origin} posited that pointing gestures originate from touch, deriving the VTL mechanism and emphasizing the significance of eye-finger position connection in gesture comprehension. Notably, Li et al.\cite{li2023understanding} proposed a touch-line transformer based on the VTL mechanism, which improved referent detection performance, validating VTL's efficacy in mitigating pointing gesture misinterpretation.

However, these studies primarily focused on distant objects. In close-proximity interactions, under gaze supervision without alignment, pointers adapt their arm posture to spatial constraints, rendering both VTL and arm-finger line mechanisms ineffective. To address these limitations, we introduce the Attention-Dynamic Touch Line for dynamic pointing gesture representation and validate its effectiveness in enhancing visual grounding performance.

\subsection{Referring expression comprehension}
Referring Expression Comprehension plays a crucial role in computer vision \cite{ji2024progressive,wang2023res}, aiming to accurately localize objects within images based on natural language descriptions. Several datasets have been developed for this task, including phrase-level (RefCOCO, RefCOCO+, RefCOCOg\cite{kazemzadeh2014referitgame}) and sentence-level (Flickr30K\cite{plummer2017flickr30k}, Visual Genome\cite{krishna2017visual}) language descriptions. Researchers have proposed various approaches to address REC. Mao et al. \cite{mao2016generation} introduced the first deep learning-based approach in this domain by developing MMI, which innovatively integrates a CNN for visual feature extraction with an LSTM for expression generation. Kamath et al.\cite{kamath2021mdetr} introduced MDETR, a transformer-based architecture that detects objects in text-conditioned images through vision-text modality fusion. Additionally, one-stage approaches\cite{yang2019fast,sadhu2019zero,luo2020multi,liao2020real}, graph-based models\cite{wang2019neighbourhood,yang2019dynamic,yang2019cross,liu2020learning}, and language pre-training models\cite{lu2019vilbert,su2019vl} have also demonstrated significant improvements.

Since the pioneering CLIP\cite{radford2021learning,zhang2024vision} study in 2021, vision-language models have achieved remarkable advancements. Numerous REC-capable methods have emerged, such as GLIP\cite{li2022grounded,zhang2022glipv2}, Grounding DINO\cite{liu2023grounding}, and DetCLIP\cite{yao2022detclip,yao2023detclipv2,yao2024detclipv3}. While these methods have advanced the field, they primarily leverage language cues, with limited incorporation of embodied nonverbal information as additional cues.

\begin{figure*}
\centering
\includegraphics[width=\textwidth]{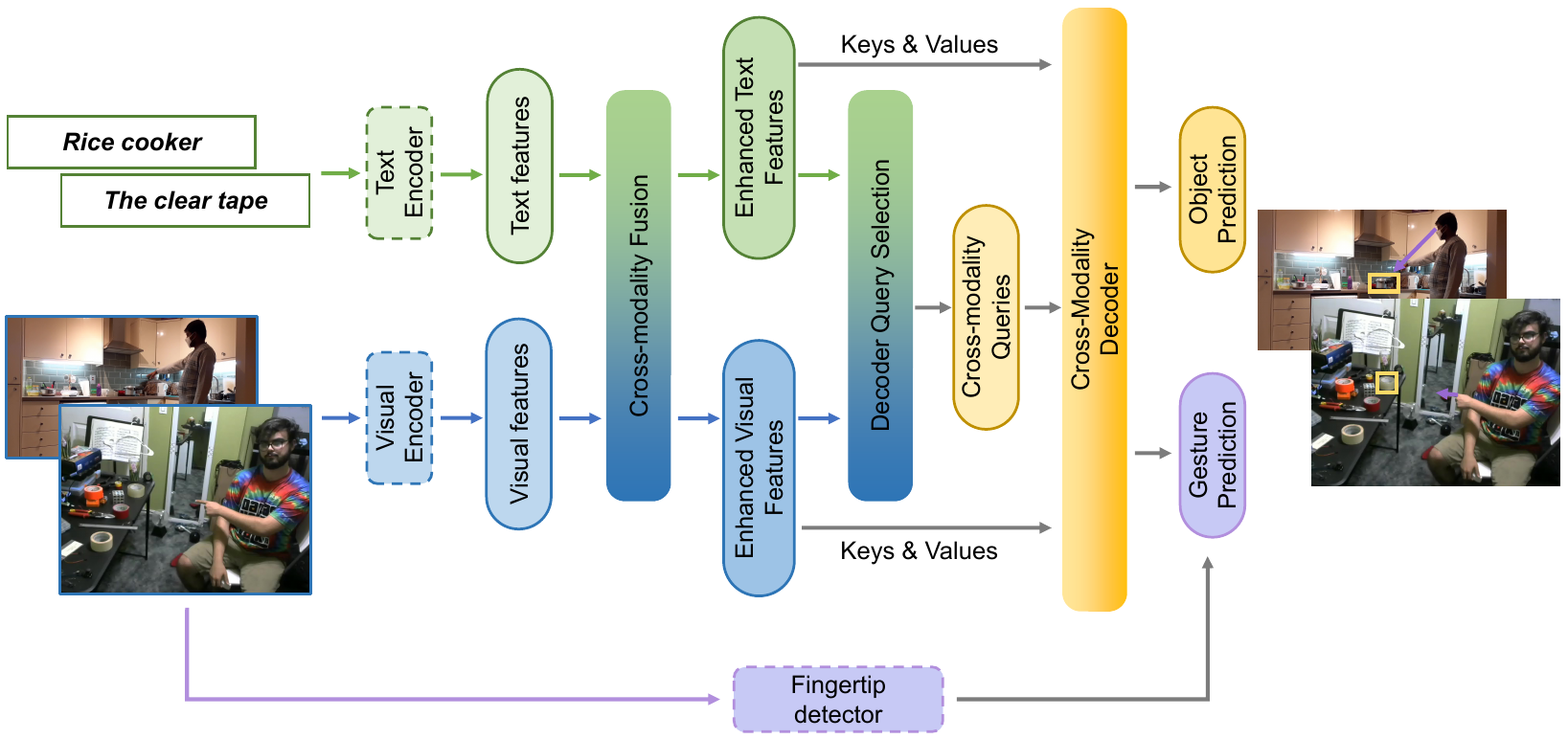}
\caption{Language and visual inputs are first encoded into initial language features and visual features, respectively. Then the two features are enhanced by cross-modality fusion. After the features are enhanced, the language-guided query selection module is used to select cross-modal queries from visual features. Finally, cross-modality decoder is used to predict the object boxes and the attention source. Attention source and the fingertip predicted by the fingertip detector jointly represent the predicted ADTL.}
\label{fig_2}
\end{figure*}

\subsection{Gesture-based Visual Grounding}
The integration of non-verbal and verbal information can significantly enhance REC performance. Human-like non-verbal cues, such as gaze or pointing gestures, provide more explicit indications to specific objects through visual understanding. Gaze-based approaches determine human visual focus on target objects. Fang et al.\cite{fang2021dual} extended gaze target detection to 3D contexts, proposing a three-stage method that predicts gaze direction, determines the field of view area and range, and locates objects within the field of view. Qian et al. \cite{qian2023gvgnet} utilized multi-modal information fusion based on text expressions, scene images, and gaze heatmaps to locate and segment target objects. While gaze alone may not provide optimal accuracy for object indication, it shows significant potential when combined with human poses.

Pointing gestures are among the most intuitive forms of human communication. Chen et al. \cite{chen2021yourefit} developed the YouRefIt dataset and benchmarked the ERU task in images and videos, employing saliency heatmaps and Part Affinity Field heatmaps to extract gesture features. Oyama et al. \cite{oyama2023exophora} combined pre-built environmental models with orientation words, object categories, and skeleton-based pointing data to locate target objects outside a robot's field of view. Lorentz et al. \cite{lorentz2023pointing} proposed a human-robot interaction process based on pointing gestures and bidirectional dialogue to guide humanoid robots in locating, picking, and placing target objects. In our study, we leverage both pointing and gaze information to refine the near-target view field, thereby improving visual grounding accuracy based on textual cues.

\section{Method}
This section presents the main design of the AD-DINO, including three key components:
\begin{enumerate}
\item{Model architecture (section III-A): Describing the implementation of target visual grounding using language and image with embodied gesture cues.}
\item{Attention-dynamic touch line Annotation (section III-B): Defining the representation of the pointing gesture based on distinct pointer-referent distances.}
\item{ Model training (section III-C): Elaborating on the specific details of the model training process.}
\end{enumerate}

\subsection{Network Architecture}
The proposed AD-DINO model, illustrated in Fig. \ref{fig_2}, comprises five main components: a visual encoder, a text encoder, a cross-modality fusion module, a query selection module, and a cross-modality decoder. The following subsections detail each component:

\subsubsection{Feature encoder and cross-modality fusion module}
\begin{figure}[h]
\centering
\includegraphics[width=3.4in]{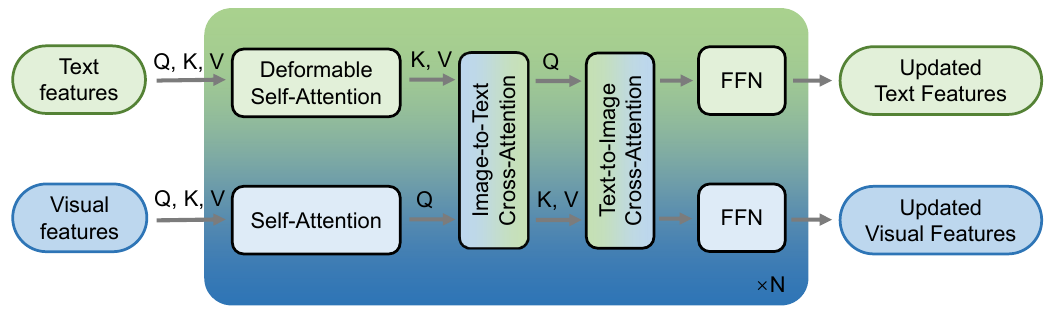}
\caption{Single layer for cross-modality fusion module.}
\label{fig_3}
\end{figure}

For each (image, text) pair, we employ the Swin Transformer as the visual encoder backbone to extract initial image features. BERT serves as the text encoder backbone to extract initial text features. We implement a cross-modality fusion module to enhance the extracted initial multimodal features, facilitating information exchange between single modality features. This module consists of multiple identical feature enhancement layers, each following a dual-feature in and dual-feature out structure as shown in Fig. \ref{fig_3} Within each layer: the text features are processed through plain self-attention, visual features are processed through deformable self-attention. Following GLIP and Grounding-DINO, we utilize sequential image-to-text and text-to-image cross-attention modules for cross-modality information fusion. The enhanced features are output through feedforward networks (FFN).

\subsubsection{Decoder Query selection module}
To select visual features most relevant to the input text as encoder queries, we implement a query selection module. This module performs dot product operations between input text features and visual features to evaluate their similarity. Image features with high similarity are selected as language-guided queries. The decoder query combines location and content components to better adapt to diverse visual features and to comprehend and express the language features of the input text.

\subsubsection{Cross-Modality Decoder Module and Finger Detector}
\begin{figure}[h]
\centering
\includegraphics[width=3.4in]{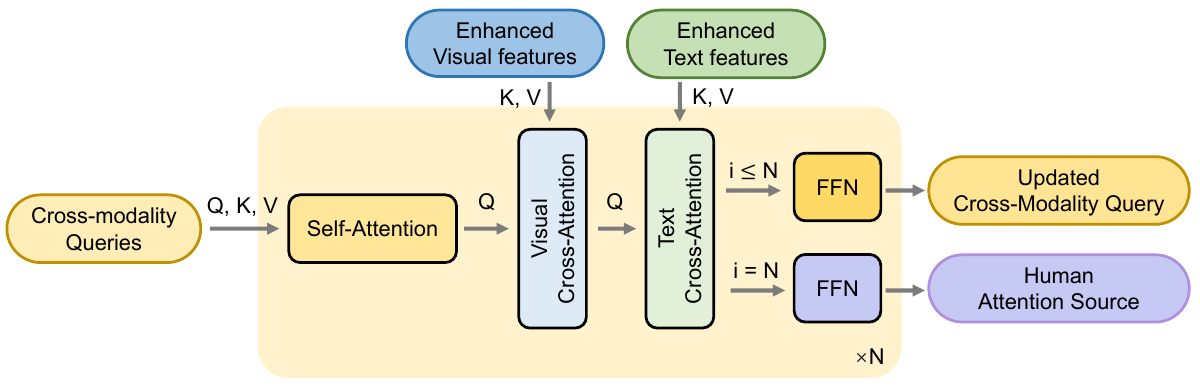}
\caption{Single layer for cross-modality decoder module.}
\label{fig_4}
\end{figure}

The cross-modality decoder, structurally similar to the fusion module, comprises multiple identical decoder layers. As depicted in Fig. \ref{fig_4}, each layer consists of a self-attention layer, a visual cross-attention layer, a text cross-attention layer and FFN layers.

For attention-dynamic touch line prediction, we divide the process into two parts: Attention source point prediction and Endpoint detection. In the final decoder layer, two separate FFNs are employed for object box prediction and attention source point prediction. The ADTL endpoint, corresponding to the pointer's fingertip, is directly detected using MediaPipe's Handmarker module. The Handmarker model parameters remain fixed during the overall training process. The predicted ADTL is then constructed as the line connecting the predicted attention source point and the detected endpoint.

\subsection{Division of Attention Source of Attention-Dynamic Touch-Line}
In our proposed approach, we define the ADTL with its endpoint at the fingertip. The attention source alternates between the eyes and the MCP, depending on the distance between the pointer and the referenced object. Specifically, we designate the eyes as the attention source when the object is approximately an arm's length away from the pointer. Conversely, for closer objects, the MCP serves as the attention source. Given the challenge of accurately determining the pointer-referent distance from a single image, we leverage natural human interaction patterns to estimate this distance. Our method evaluates the spatial relationship using three key segments of the upper limb: the index finger (IF), forearm (FA), and upper arm (UA). We take these segments as three vectors: $\overrightarrow{IF} = (x_f–x_{MCP}, y_f-y_{MCP})$, $\overrightarrow{FA} = (x_w–x_e, y_w-y_e)$, $\overrightarrow{UA} = (x_e–x_s, y_e-y_s)$,where $(x_f, y_f)$, $(x_{MCP}, y_{MCP})$, $(x_w, y_w)$, $(x_e, y_e)$, $(x_s, y_s)$ represents the coordinates of fingertip, MCP, wrist, elbow, and shoulder. We use MediaPipe as the human bone detection model, to detect these joints. 

To evaluate the integral directional consistency level of three vectors, as shown in equation (1), taking the two segment vectors corresponding to the two maximum cosine similarity, denoted as $\overrightarrow{S}_{sum1}$ and $\overrightarrow{S}_{sum2}$, respectively.

\begin{equation}
\begin{split}
    \max(&cos\_sim_{IF\_FR}, cos\_sim_{FR\_UA}, cos\_sim_{IF\_UA}) \\
    &\rightarrow (\overrightarrow{S}_{sum1}, \overrightarrow{S}_{sum2})
\end{split}
\end{equation}
Where \textit{cos\_sim} represents the cosine similarity of two vectors.

Sum vector $\overrightarrow{S}_{sum} = \overrightarrow{S}_{sum1} + \overrightarrow{S}_{sum2}$. The remaining segment vector is denoted as $\overrightarrow{S}_{rem}$. Then calculating the cosine similarity between the $\overrightarrow{S}_{sum}$ and $\overrightarrow{S}_{rem}$.As depicted in equation (2), when $cos\_sim_{sum\_rem}$ is greater than the threshold, we consider pointer to have extended his arm and the object distance pointer exceeds the length of his arm, attention source is positioned as eye; Conversely, we consider pointer to have bent his arm and the distance between the object and pointer is less than the length of his arm, attention source is positioned as MCP.

\begin{equation}
AS = \begin{cases}
eye,&{\text{if}}\ cos\_sim_{sum\_rem}>t \\ 
{MCP,}&{\text{otherwise.}} 
\end{cases}
\end{equation}
Where \textit{AS} represents attention source, \textit{t} represents the threshold, \textit{t} = 0.95.

\subsection{Model Training}
\subsubsection{Explicit Learning of Pointing Gesture}
In general, to provide the observer a clear indication of the referent, the pointer's source of attention is typically highly collinear with the fingertip and the object. We assess the alignment of these points using cosine similarity:

\begin{equation}
AE = cos\_sim[(x_o-x_a,y_o-y_a),(x_o-x_f,y_o-y_f)]
\end{equation}
Where, \textit{AE} represents evaluation of the alignment degree of three points. the $(x_a,y_a)$,$(x_o,y_o)$ respectively represent the coordinates of attention source and the bounding box center of the object.

To achieve the unity of referent localization and alignment with the pointing gesture representation line in ERU tasks, the alignment evaluation loss function is defined as shown in equation (4):

\begin{equation}
L_{AE} = ReLU(AE_{gt}-AE_{pred})
\end{equation}
Where, $AE_{gt}$ is calculated from the bounding box of the ground truth, and $AE_{pred}$ is calculated from the predicted bounding box. In both calculations, the source of attention is derived from the ground truth.
\subsubsection{Loss Function}
For the total loss function during training process, the specific expression is as equation (5):

\begin{equation}
L = L_{box}+L_{GIoU}+L_{alignment} +L_{AS}+L_{AE}
\end{equation}
Where $L_{box}$ and $L_{GIoU}$ are L1 loss and GIOU loss for bounding box regression respectively. $L_{alignment}$ is contrastive focal loss between regional visual features and textual token as in \cite{liu2023grounding}. For ADTL, the $L_{AS}$ is the L1 loss for attention source regression. The $L_{AE}$ is detailed above in section III-B.

\begin{table*}[!t]
\caption{A comparison of state-of-the-art methods in ERU\label{tab:table1}}
\centering
\begin{tabular}{cccc}
\hline
Model & IoU=0.25 & IoU=0.5 & IoU=0.75\\
\hline
FAOA\cite{yang2019fast} & 44.5 & 30.4 & 8.5\\
ReSC\cite{yang2020improving}&	49.2&	34.9	&10.5\\
GLIP-T\cite{zhang2022glipv2}	&50.3&	40.0&	29.3\\
GLIP-L\cite{zhang2022glipv2}	&49.4	&46.1	&35.8\\
Grounding-DINO-T\cite{liu2023grounding} &46.6 &44.1 &34.8\\
Grounding-DINO-L\cite{liu2023grounding} &51.2 &47.6 &36.3\\
$\text{YouRefIt}_\text{PAFonly}$ \cite{chen2021yourefit} &52.6 &	37.6 &12.7\\
$\text{YouRefIt}_\text{Full}$ \cite{chen2021yourefit} &54.7	&40.5	&14.0\\
REP\cite{shi2022spatial}	&58.8	&45.7	&18.8\\
TOUCH-IN-LINE  \cite{li2023understanding} & 69.5 & 60.7 & 35.5\\
TOUCH-IN-LINE (VTL) \cite{li2023understanding} & 71.1 & 63.5 &	39.0\\
TOUCH-IN-LINE (ADTL) \cite{li2023understanding} & 72.2 & 65.0 & 39.3\\
\hline
Ours(VTL) & 75.1 & 71.6 & 53.7\\
Ours(FL) & 74.0 & 70.1 & 52.4\\
Ours(ADTL) & \textbf{76.3} & \textbf{72.4} & \textbf{55.4}\\
\hline
Human & 94.2 & 85.8 & 53.3\\
\hline

\end{tabular}
\end{table*}

\subsubsection{Implicit Learning of Pointing Gesture}
To assess our model's implicit learning of pointing gesture information from images, we conducted an ablation study. This experiment involved modifying both the input images and the model architecture. For image modification, we eliminated human figures from the images and inpainted the removed areas based on the surrounding background, employing the MAT \cite{li2022mat}. Concurrently, we adjusted the model by removing the FFN responsible for predicting the human attention source, retaining only the object prediction output. This experimental setup allows us to isolate the impact of explicit human pose information on our model's performance. By comparing the results of this modified setup with our original model, we can evaluate the extent to which our model relies on implicit gesture cues versus explicit human pose information. Detailed results and analysis of this ablation study are presented in section IV-D.

\section{Experiments}
\subsection{Dataset and Evaluation Metrics}
This study utilizes the YouRefIt dataset, an embodied reference dataset where agents employ both language and gestures to refer to other agents in shared physical environments. While the dataset includes both videos and images, our study focuses exclusively on image data. The training set comprises 2950 samples, with 1245 samples in the test set. We retain the original text annotations from the YouRefIt dataset and augment it with FL annotations. VTL annotations follow \cite{li2023understanding}, while ADTL annotations are selected from FL and VTL references, as detailed in section III-B.

To facilitate comparison with previous methods, we adopt the experimental setup from \cite{chen2021yourefit} and \cite{li2023understanding}, employing three Intersection over Union (IoU) thresholds for accuracy reporting: 0.25, 0.5, and 0.75. A prediction is deemed correct if the calculated IoU between the predicted bounding box and the ground truth exceeds the set threshold. Additionally, we report Generalized Intersection over Union (GIoU) results to provide a broader evaluation perspective, as discussed in section IV-C.

\begin{figure*}[h]
\centering
\subfloat[]{%
    \includegraphics[width=0.31\textwidth]{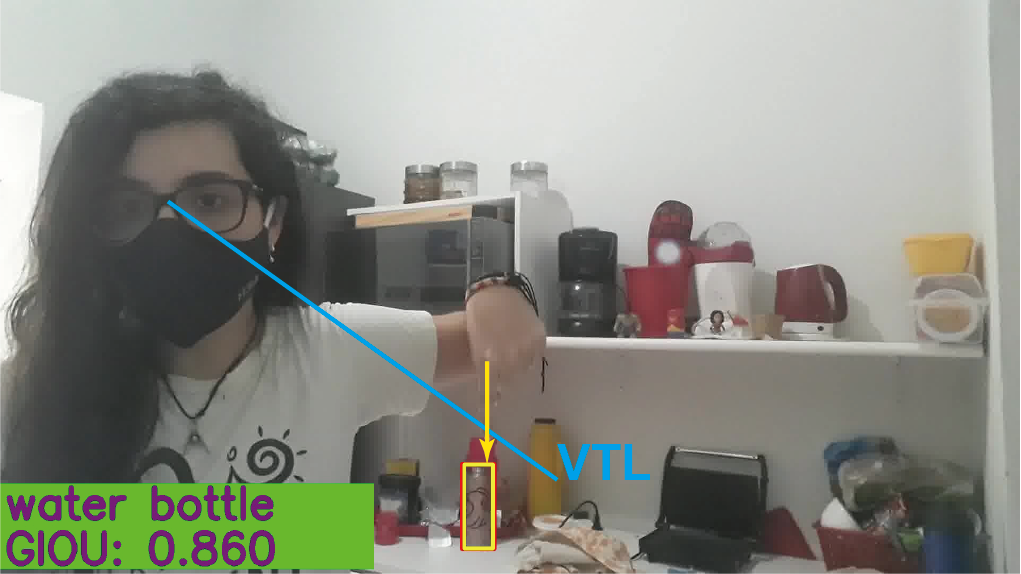}
}
\hfill
\subfloat[]{%
    \includegraphics[width=0.31\textwidth]{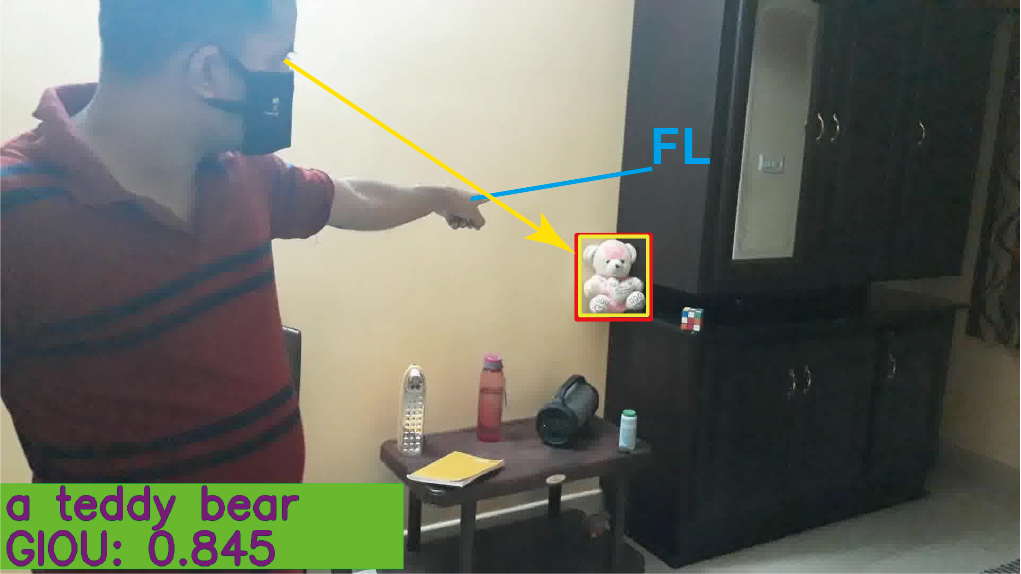}
}
\hfill
\subfloat[]{%
    \includegraphics[width=0.31\textwidth]{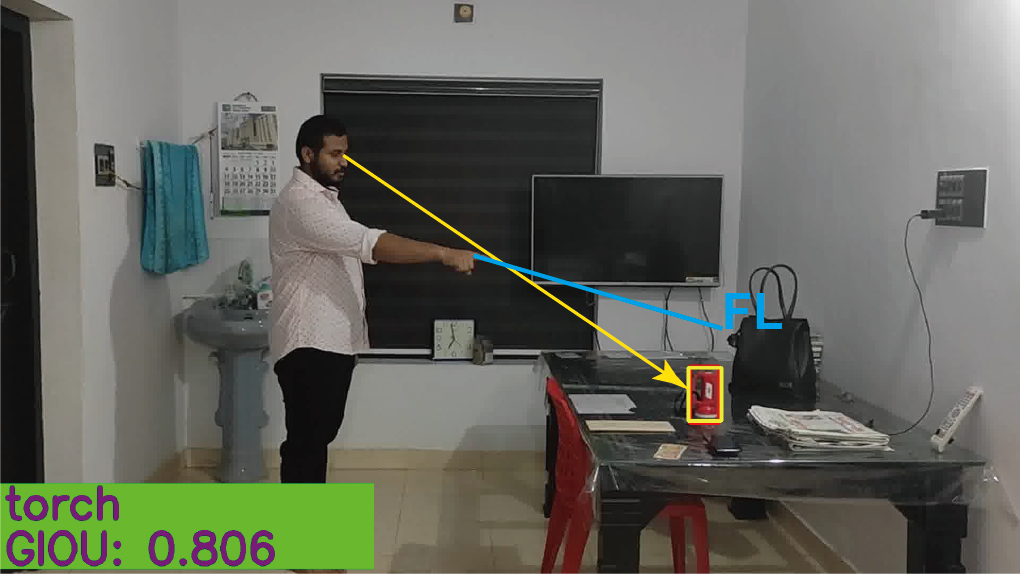}
}

\vspace{0.01in}

\subfloat[]{%
    \includegraphics[width=0.31\textwidth]{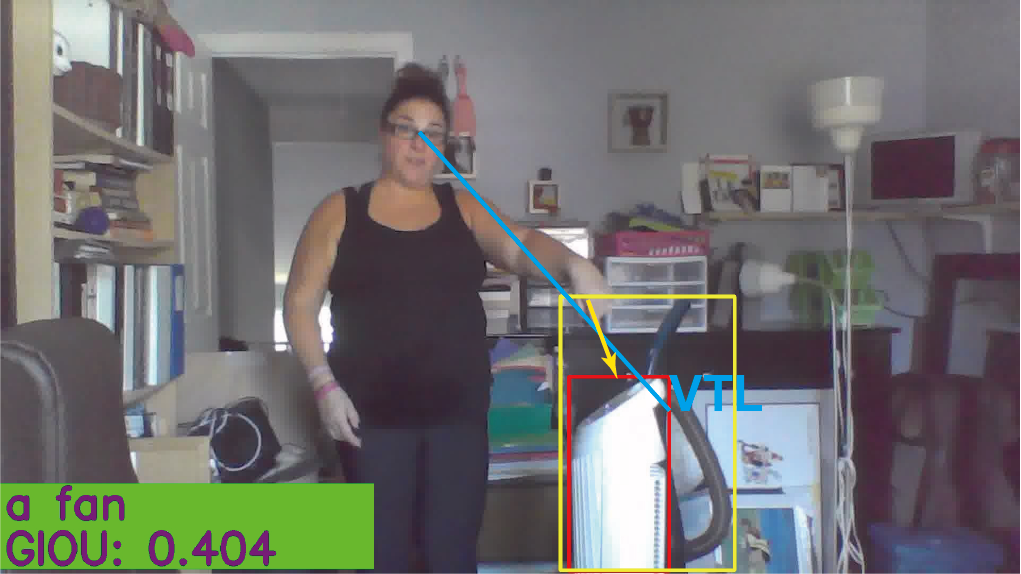}
}
\hfill
\subfloat[]{%
    \includegraphics[width=0.31\textwidth]{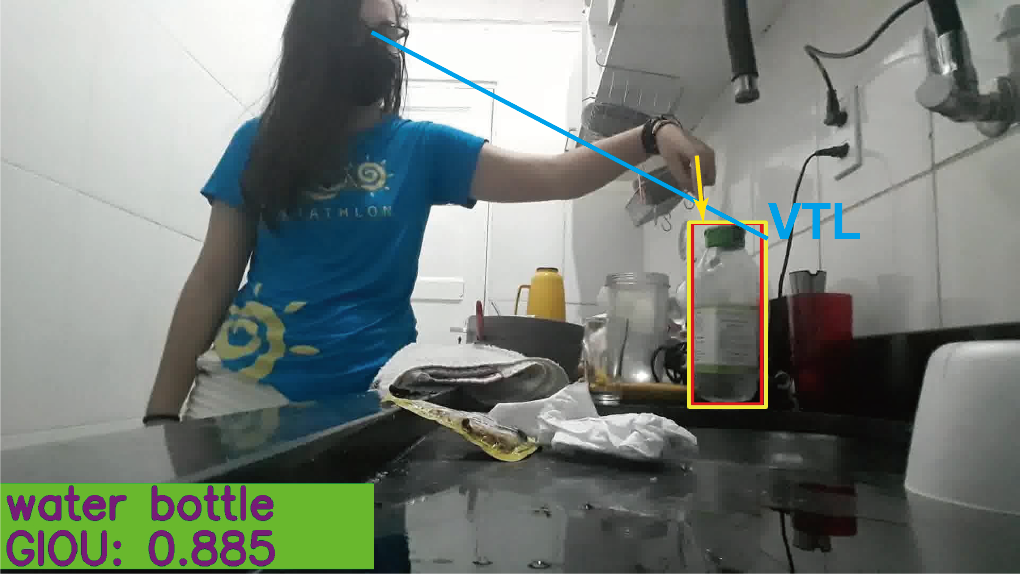}
}
\hfill
\subfloat[]{%
    \includegraphics[width=0.31\textwidth]{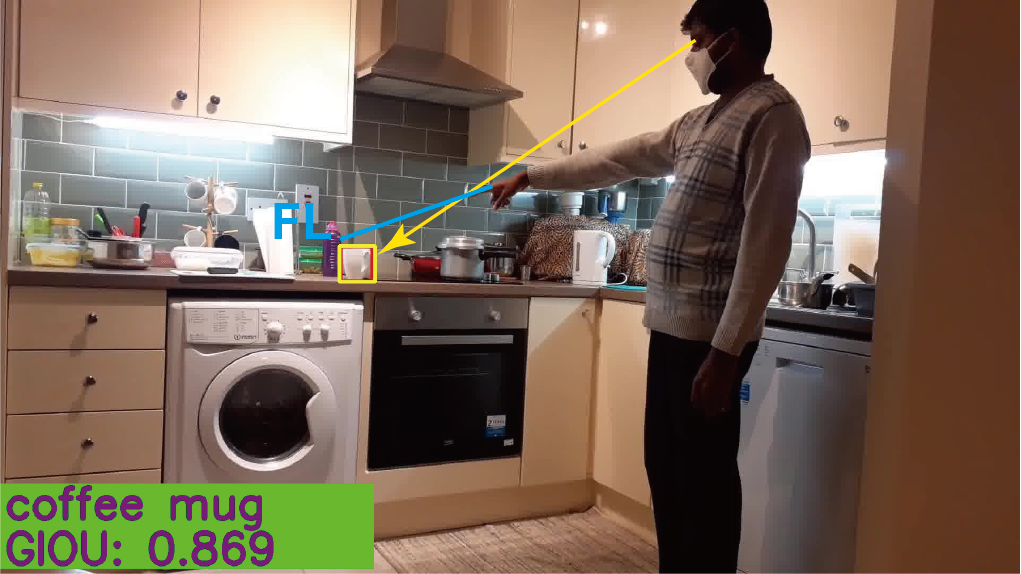}
}

\caption{The demonstration of distance-aware character of ADTL. Relative to distance-unaware VPT methods, ADTL can establish a more accurate alignment with referent based on interaction distance awareness. The ADTLs in (a, d, e) take the form of FL, while those in (b, c, f) take the form of VTL. Yellow and red rectangles represent the ground truth and ADTL predicted bounding boxes, respectively. Yellow arrows indicate the ADTLs, and blue lines represent the interpretation mode abandoned by the ADTL. The words in green on the first line are the natural language input, and those on the second line show the GIOU value between the ground truth bounding box and the predicted bounding box.}
\label{fig_5}
\end{figure*}

\begin{figure*}[!t]
\centering
\begin{tabular}{ccc}
    \textbf{VTL} & \textbf{FL} & \textbf{NEGKP} \\[0.1em]
    \includegraphics[width=0.2\textwidth]{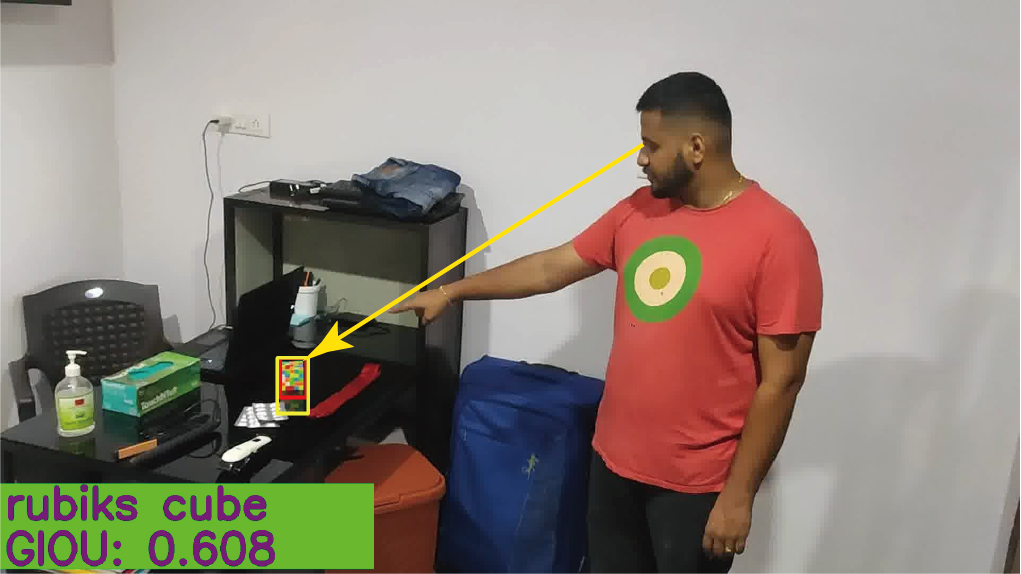} &
    \includegraphics[width=0.2\textwidth]{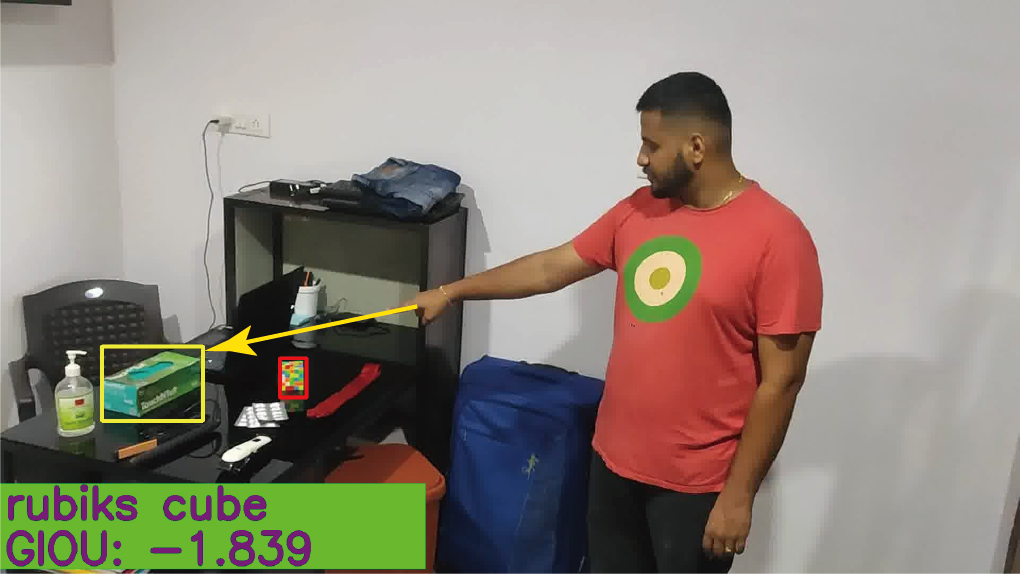} &
    \includegraphics[width=0.2\textwidth]{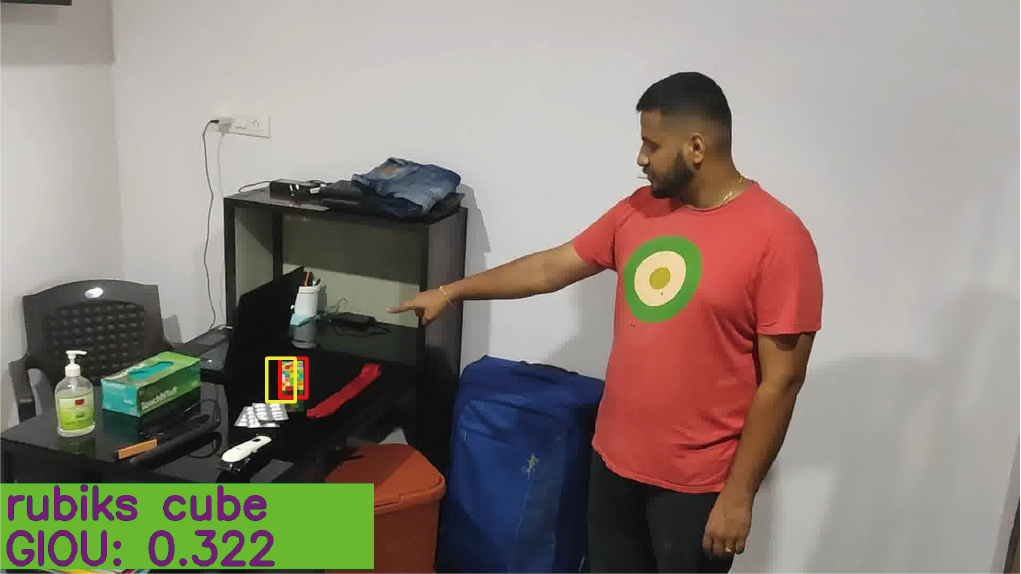} \\
    \multicolumn{3}{c}{(a)} \\[0.1em]
    \includegraphics[width=0.2\textwidth]{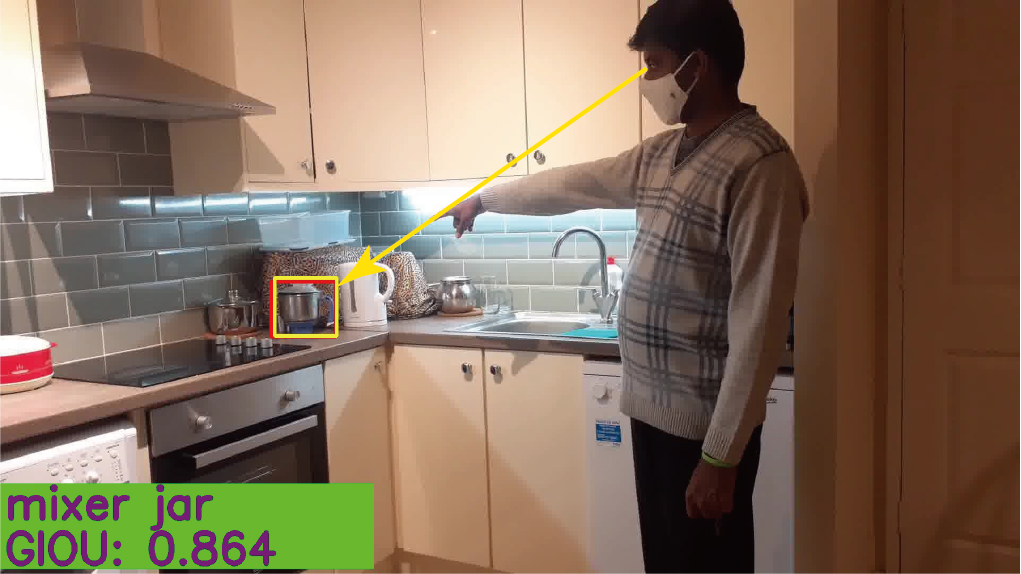} &
    \includegraphics[width=0.2\textwidth]{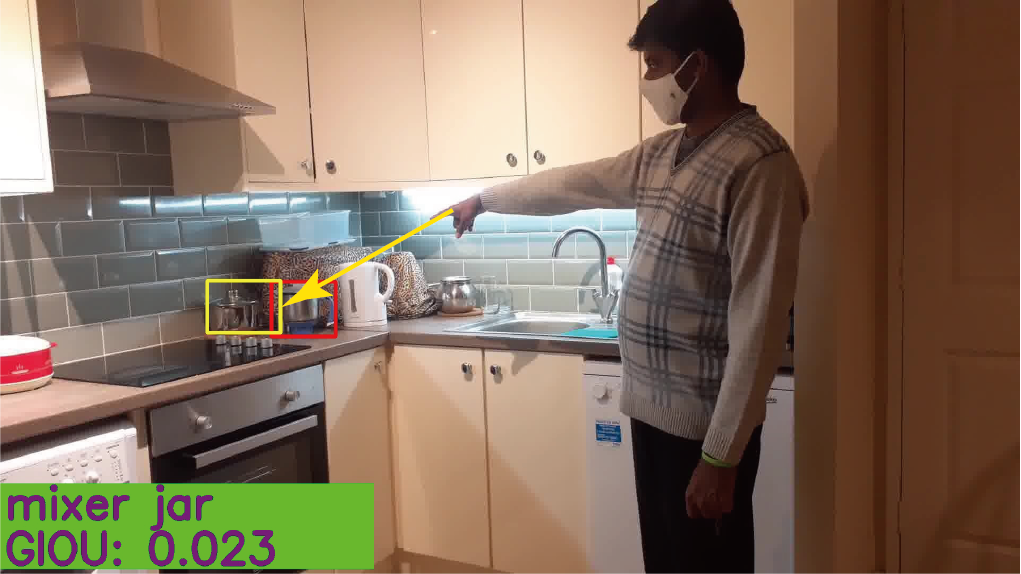} &
    \includegraphics[width=0.2\textwidth]{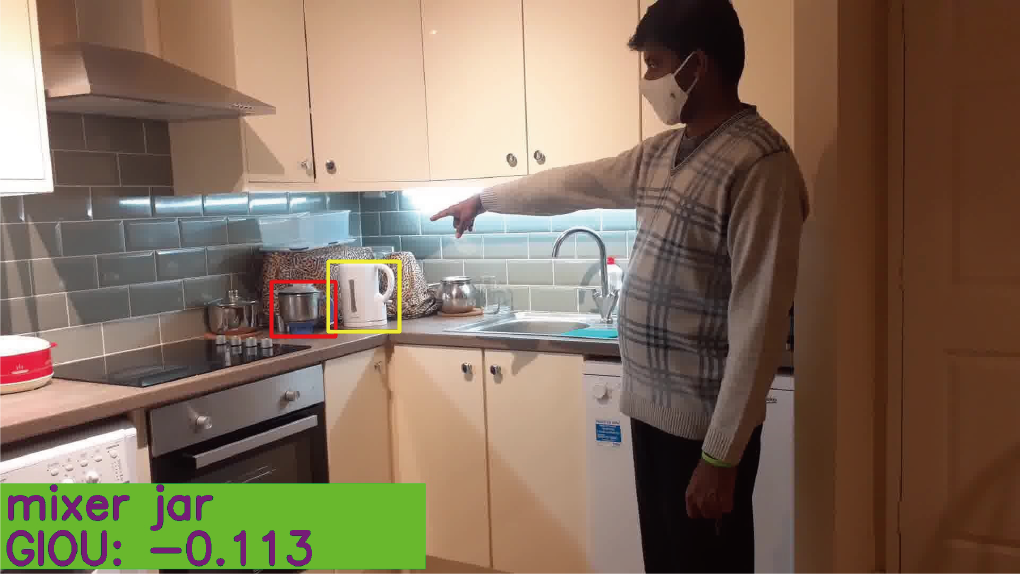} \\
    \multicolumn{3}{c}{(b)}\\[0.1em]
    \includegraphics[width=0.2\textwidth]{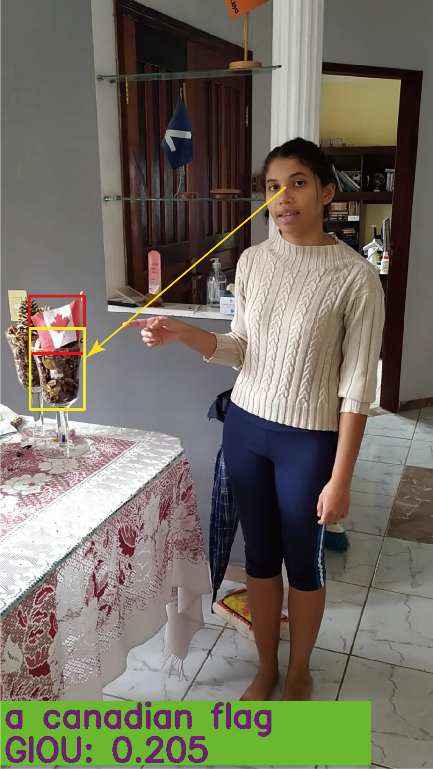} &
    \includegraphics[width=0.2\textwidth]{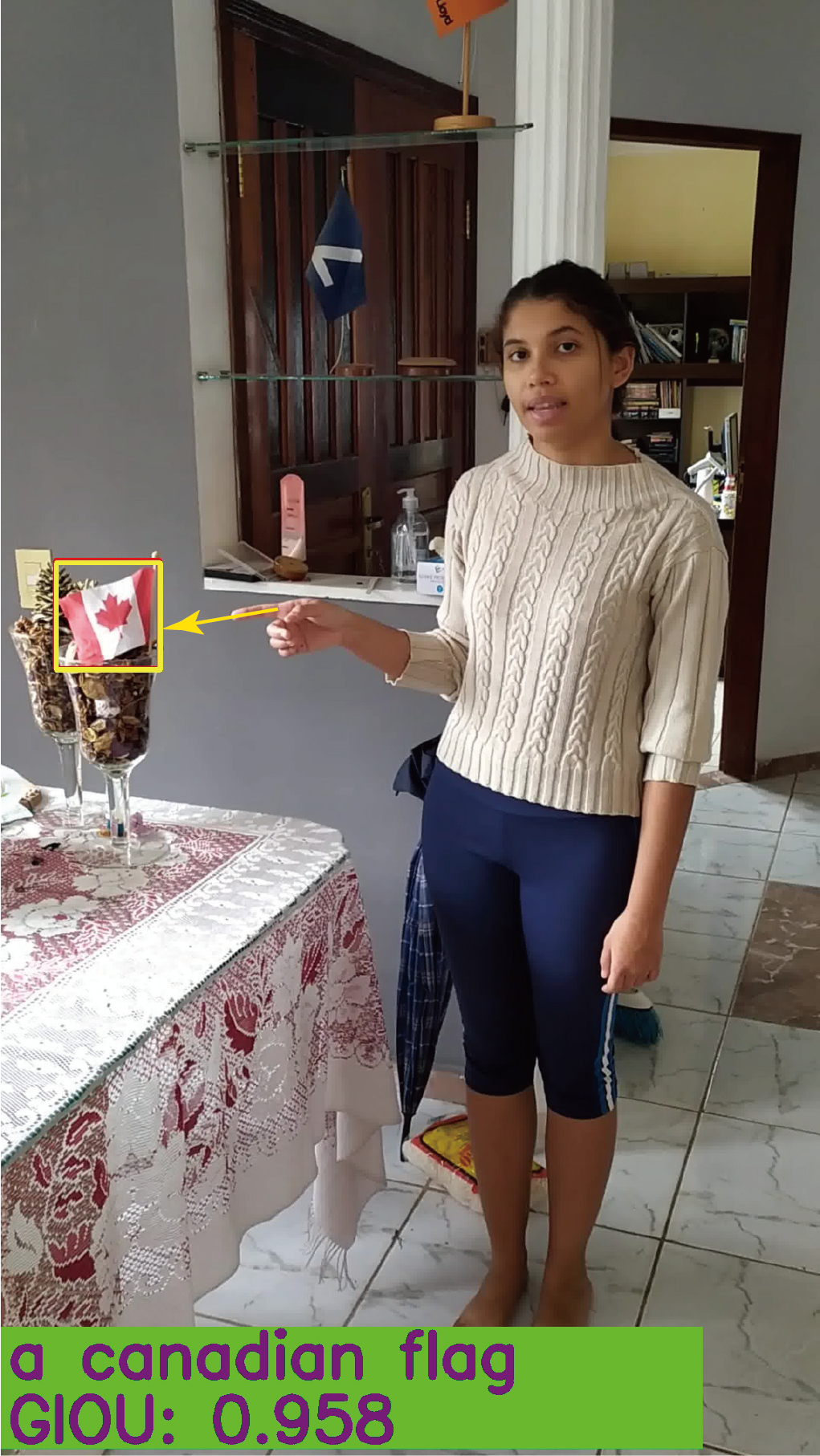} &
    \includegraphics[width=0.2\textwidth]{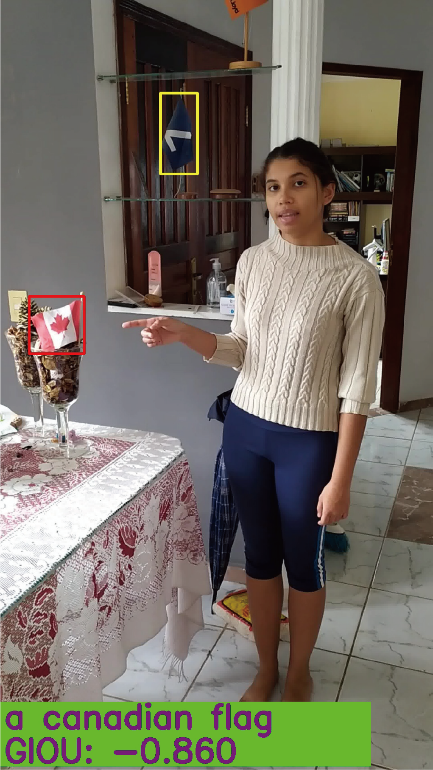} \\
    \multicolumn{3}{c}{(c)}\\[0.1em]
    \includegraphics[width=0.2\textwidth]{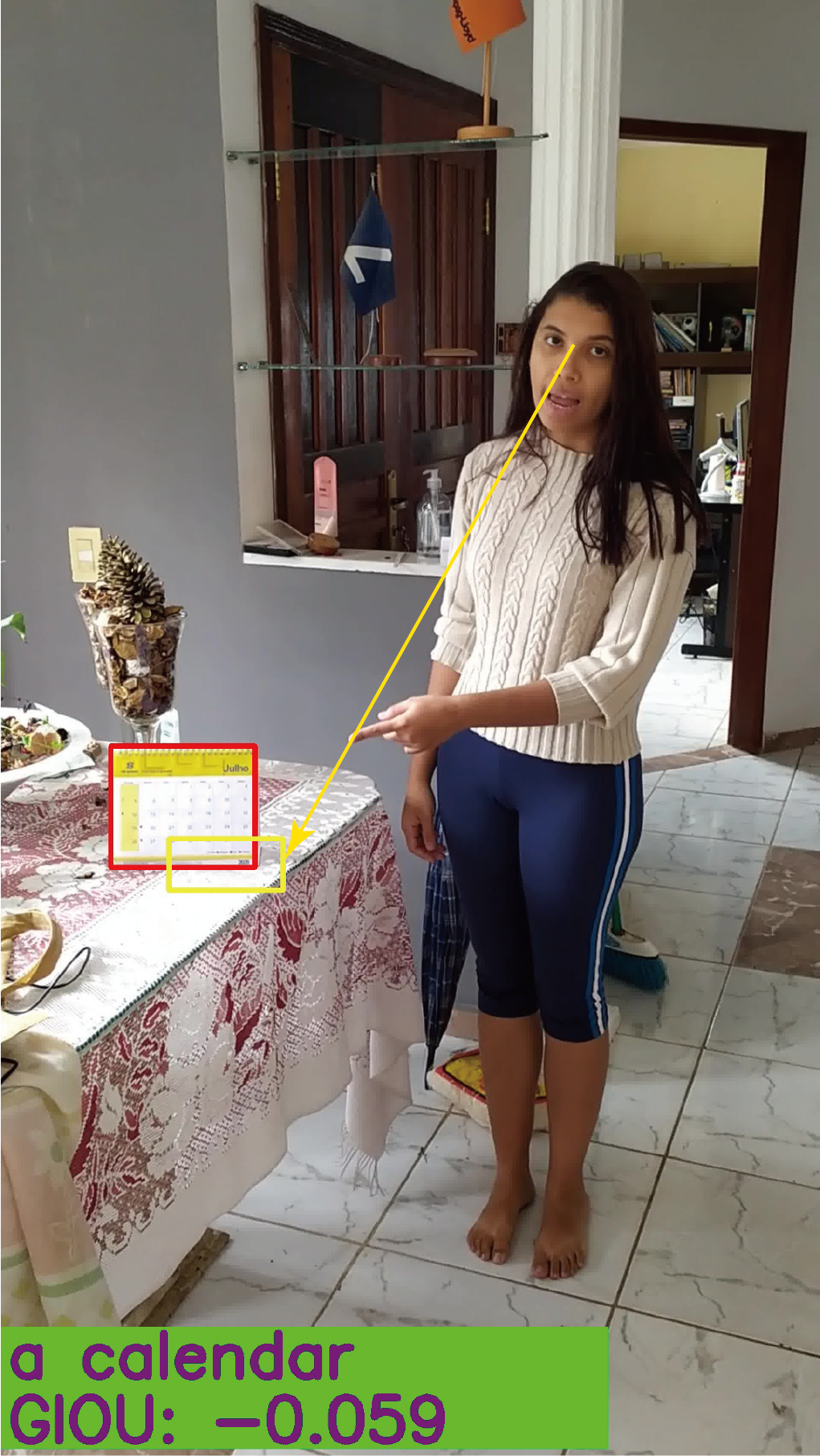} &
    \includegraphics[width=0.2\textwidth]{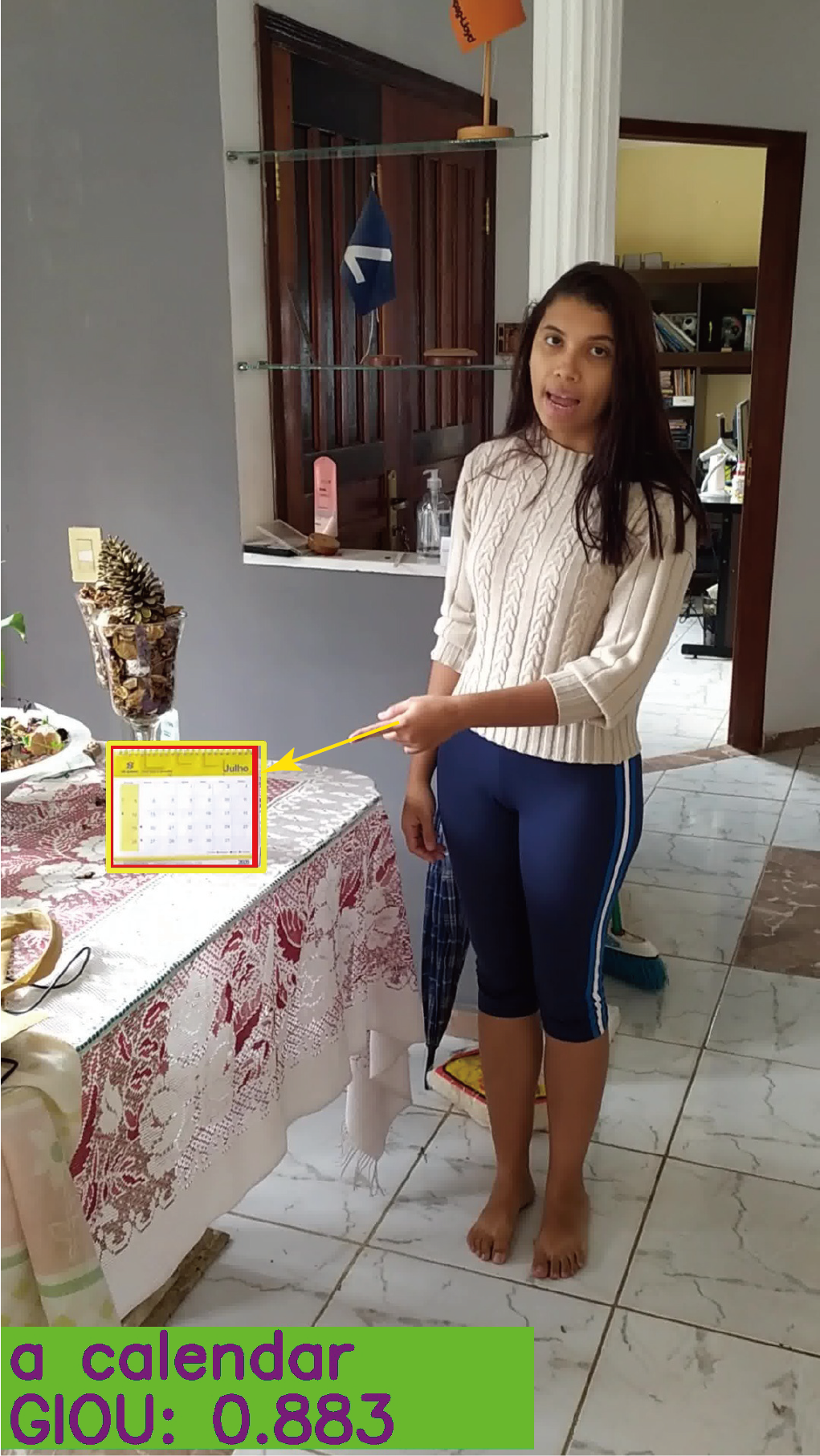} &
    \includegraphics[width=0.2\textwidth]{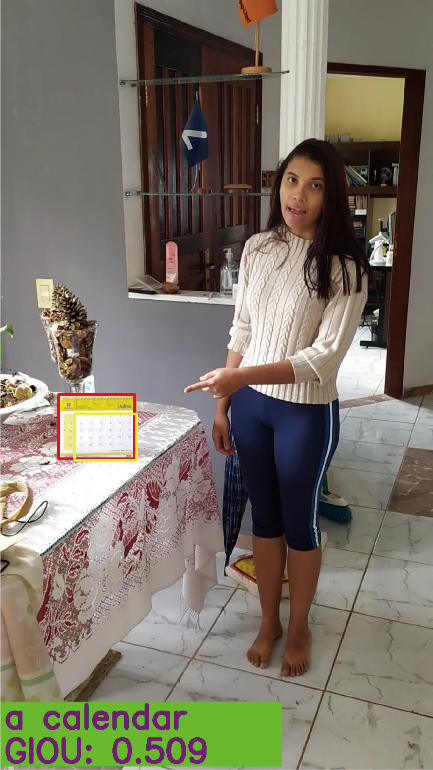} \\
    \multicolumn{3}{c}{(d)}\\[0.1em]
\end{tabular}
\caption{Comparison of explicit learning methods of pointing gesture. Model with explicit learning methods increase the referent locating accuracy, relative to method without explicit gestural key points. Images in the right column is the results with no explicit gestural key points. Yellow arrows in left column represent the VTL, in middle column represent the FL. Yellow and red rectangles represent the ADTL predicted and ground truth bounding boxes, respectively. The words in green on the first line are the natural language input, and those on the second line show the GIOU value between the ground truth bounding box and the predicted bounding box.}
\label{fig_6}
\end{figure*}
%%%%%%%%%%%%%%%%%%%%%%%%%%%%%%%%%%%%%%%%%%%%%%

\begin{figure*}[!t]
\centering
\includegraphics[width=0.8\textwidth]{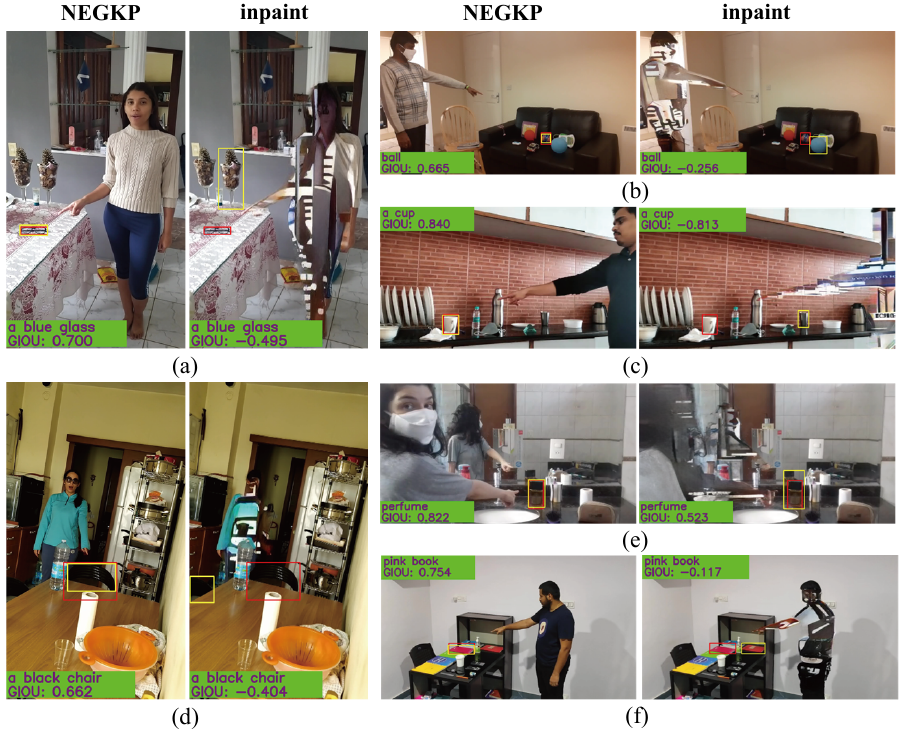}
\caption{Comparison of object localization performance with different methods. The column labeled 'NEGKP' illustrates the model's performance without explicit gesture key points. Columns labled 'inpaint' demonstrate the model's performance when human subjects are removed from the image and the background is subsequently inpainted. }
\label{fig_7}
\end{figure*}

%%%%%%%%%%%%%%%%%%%%%%%%%%%%%%%%%%%%%%%%%%%%%%%

\subsection{Comparison of ERU Performance with State-of-the-art Methods}
TABLE \ref{tab:table1} demonstrates that our ADTL-based model achieves new SOTA performance across all three IoU thresholds, with accuracies of 76.3\%, 72.4\%, and 55.4\%, respectively. Notably, at the 0.75 IoU threshold, our model with ADTL shows substantial improvement, surpassing the YouRefIt benchmark by 41.4\% and the previous SOTA method\cite{li2023understanding} by 16.4\%. Significantly, both ADTL and VTL methods outperform human performance under this threshold for the first time. At the 0.5 IoU threshold, our model outperforms the SOTA model by 7.4\% and 8.1\% with ADTL and VTL, respectively.

For reference, we also evaluate SOTA visual grounding models \cite{yang2019fast,yang2020improving,zhang2022glipv2,liu2023grounding}, employing both Swin-T and Swin-L backbones in Grounding-DINO and GLIP. These models are applied to YouRefIt as referring expression comprehension tasks, without explicitly utilizing nonverbal signals in the images.

\subsection{GIoU Performance and DA-VPT Effectiveness Evaluation}
To validate our model's visual grounding capabilities and the significance of DA-VPT, we present GIoU performance metrics in TABLE \ref{tab:table2}, using thresholds consistent with IoU. Comparing TABLE \ref{tab:table1} and TABLE \ref{tab:table2}, the advantages of the ADTL method over distance-unaware VPT gesture understanding are evident. ADTL consistently outperforms VTL and FL across all evaluation metrics and thresholds, with VTL surpassing FL. Under the 0.25 IoU threshold, ADTL demonstrates improvements of 0.9\% and 2.2\% over VTL and FL, respectively.

\begin{table*}[!b]
\caption{A Comparison of GIoU performance of different gesture representation methods\label{tab:table2}}
\centering
\begin{tabular}{ccccc}
\hline
Property &Method & GIoU=0.25 & GIoU=0.5 & GIoU=0.75\\
\hline
\multirow{2}{*}{\shortstack{Distance-unaware VPT}} &Ours(VTL) &74.7 &71.5 &52.7\\
                                      &Ours(FL) &73.4 &69.8	&51.2\\
\hline                                      
Distance-aware VPT   &Ours(ADTL) &75.6 &72.0 &54.9\\

\hline
\end{tabular}
\end{table*}

In close-range interactions where the distance is less than arm's length, the slightly inferior performance of VTL relative to ADTL can be attributed to the unnecessary requirement for eye-finger-object alignment. Under gaze supervision, the pointer can manipulate the ADTL to accurately intersect the predicted object's spatial position in FL form. The flexibility of elbow, wrist, and MCP joints allows for arbitrary FL gesture adjustments while maintaining object targeting. In such scenarios, VTL may deviate significantly from the object's center or miss the target bounding box entirely, impeding the model's ability to leverage gesture information for visual grounding. Fig. \ref{fig_5}(a) illustrates this issue, where VTL fails to intersect the target water bottle's bounding box, potentially confusing embodied reference and language cues during prediction.

Conversely, in long-range interactions where the distance is beyond arm's length, pointers typically extend their arms to achieve eye-finger-object alignment, allowing ADTL in VTL form to intersect the target object. Fig. \ref{fig_5}(b) demonstrates VTL's success in matching language cues, while FL fails. In both scenarios, higher alignment between embodied reference and language cues enhances the model's prediction accuracy. ADTL's dynamic understanding of embodied references, free from single-mechanism constraints, significantly reduces pointing gesture misinterpretations.

To further validate the DA-VPT-based ADTL method's efficacy in pointing gesture understanding, we apply it to the SOTA TOUCH-IN-LINE model. Results in TABLE \ref{tab:table1} show that TOUCH-IN-LINE with ADTL outperforms that with VTL by 1.1\%, 1.5\%, and 0.3\% at 0.25, 0.5, and 0.75 IoU thresholds, respectively.

\subsection{Explicit Learning of Pointing Gesture}
To assess the impact of explicit gesture learning on our model's performance, we compare models with explicit gesture information (ADTL, VTL, and FL) against those with no explicit gestural key points (NEGKP). TABLE \ref{tab:table3} demonstrates that ADTL and VTL models consistently outperform models with NEGKP across all IoU thresholds. This performance advantage may be attributed to factors similar to those discussed in section IV-C. The addition of highly aligned embodied references enhances the model's prediction accuracy compared to language cues alone, as evidenced by ADTL and VTL results, as well as FL performance at 0.25 and 0.5 IoU thresholds.

\begin{table}[!htbp]
\caption{A Comparison of Different Gesture Representation in Explicit Learning\label{tab:table3}}
\centering
\begin{tabular}{cccc}
\hline
Method & IoU=0.25 & IoU=0.5 & IoU=0.75\\
\hline
Ours(VTL) &75.1 &71.6 &53.7\\
Ours(FL) &74.0 &70.1 &52.4\\                                    
Ours(ADTL) &76.3 &72.4 &55.4\\
Ours(NEGKP) &73.6 &69.5 &52.3\\
\hline
\end{tabular}
\end{table}

However, under the 0.75 IoU threshold, the FL model's advantage over models without explicit gestural key points becomes barely perceptible. This observation may be due to: (1) The strict evaluation threshold and FL's unreliability in orientation indication. (2) At lower thresholds (0.25 and 0.5), the approximate direction provided by FL suffices to define the spatial scope for object location. Fig. \ref{fig_6} illustrates the performance of explicit gesture learning.

\subsection{Implicit Learning of Pointing Gesture}
We explore our model's capacity for implicit learning of embodied references through two sets of comparison tests. The first set employs original images without gesture annotations (eye, MCP, wrist, elbow, shoulder). The second set uses images with humans removed and inpainted based on surrounding backgrounds. In both sets, attention source prediction is eliminated, and IoU thresholds remain consistent with previous experiments. TABLE \ref{tab:table4} presents the specific results. Despite the absence of artificial gesture annotations, the lack of visual embodied reference information in the images leads to varying degrees of performance decline across all IoU thresholds. This observation aligns with findings reported in TOUCH-IN-LINE.

\begin{table}[!htbp]
\caption{Effects of Implicit Learned Pointing Gesture\label{tab:table4}}
\centering
\begin{tabular}{cccc}
\hline
Method & IoU=0.25 & IoU=0.5 & IoU=0.75\\
\hline
Ours(NEGKP) &73.6 &69.5 &52.3\\
Ours(inpainting) &71.8 &68.6 &52.4\\
\hline
\end{tabular}
\end{table}

The performance decline resulting from human figure removal indicates that human gestures contain significant directional information in visual features. The visual grounding task in inpainted images remains similar to general REC tasks. The presence of humans with referring gestures in images implicitly provides additional visual cues, which can offer more accurate orientation information when multiple objects match verbal cue descriptions. Fig. \ref{fig_7}(b) illustrates this scenario, where multiple balls are present in the image. Without additional positional cues, the larger, clearer ball on the right is more likely to be successfully located, despite the target being the smaller ball in the middle. Thus, gesture information aids in visual referent identification, reducing the need for complex supplementary language descriptions.

Interestingly, at the 0.75 IoU threshold, performance on inpainted images surpasses that on original images. This phenomenon may be attributed to human presence acting as a visual distraction, particularly when precise localization is required. The complex features of human figures may cause the model to confuse boundaries between the person and the target region, affecting precise localization. Conversely, the more consistent background of inpainted images reduces distractions, allowing the model to focus more effectively on precise target localization.

\subsection{Effects of Object Sizes}
TABLE \ref{tab:table5} presents our analysis of object size influence on visual grounding success rates. Following \cite{chen2021yourefit}, we categorize object sizes into three groups. Consistent with other SOTA methods and human performance, our method demonstrates higher success rates for larger objects compared to smaller ones. Two factors may contribute to the decline in model performance as object size decreases: (1) Reduced probability of the touch line intersecting smaller objects, coupled with amplified relative deviation from the object's center. (2) Limited image features due to small object areas in images, constrained by picture quality and resolution, leading to reduced performance success rates.

\begin{table*}[!t]
\caption{Performance comparison of different methods on the YouRefIt dataset\label{tab:table5}}
\centering
\resizebox{\textwidth}{!}{%
\begin{tabular}{l|cccc|cccc|cccc}
\hline
\multirow{2}{*}{Method} & \multicolumn{4}{c|}{IoU = 0.25} & \multicolumn{4}{c|}{IoU = 0.5} & \multicolumn{4}{c}{IoU = 0.75} \\
\cline{2-13}
 & All & S & M & L & All & S & M & L & All & S & M & L \\
\hline
FAOA  & 44.5 & 30.6 & 48.6 & 54.1 & 30.4 & 15.8 & 36.2 & 39.3 & 8.5 & 1.4 & 9.6 & 14.4 \\
ReSC  & 49.2 & 32.3 & 54.7 & 60.1 & 34.9 & 14.1 & 42.5 & 47.7 & 10.5 & 0.2 & 10.6 & 20.1 \\
$\text{YouRefIt}_\text{PAFonly}$ & 52.6 & 35.9 & 60.5 & 61.4 & 37.6 & 14.6 & 49.1 & 49.1 & 12.7 & 1.0 & 16.5 & 20.5 \\
$\text{YouRefIt}_\text{Full}$ & 54.7 & 38.5 & 64.1 & 61.6 & 40.5 & 16.3 & 54.4 & 51.1 & 14.0 & 1.2 & 17.2 & 23.3 \\
TOUCH-IN-LINE (EWL) & 69.5 & 56.6 & 71.7 & 80.0 & 60.7 & 44.4 & 66.2 & 71.2 & 35.5 & 11.8 & 38.9 & 55.0 \\
TOUCH-IN-LINE (VTL) & 71.1 & 55.9 & 75.5 & 81.7 & 63.5 & 47.0 & 70.2 & 73.1 & 39.0 & 13.4 & 45.2 & 57.8 \\
TOUCH-IN-LINE (ADTL) & 72.2 & 56.6 & 76.8 & 82.9 & 65.0 & 48.2 & 74.0 & 72.9 & 39.3 & 13.5 & 45.5 & 58.3 \\
\hline
Ours (NEGKP) & 73.6 & 65.2 & 74.3 & 81.2 & 69.5 & 60.2 & 71.9 & 76.3 & 52.3 & 36.6 & 56.4 & 63.8 \\
Ours (inpainting) & 71.8 & 60.0 & 74.6 & 81.4 & 68.6 & 56.0 & 72.7 & 77.1 & 52.4 & 34.6 & 57.0 & 65.7 \\
Ours (VTL) & 75.1 & 64.9 & 78.1 & 82.4 & 71.6 & 61.2 & 75.7 & 78.0 & 53.7 & 33.3 & 60.2 & 67.6 \\
Ours (FL) & 74.0 & 65.2 & 76.5 & 80.2 & 70.1 & 60.5 & 74.3 & 75.6 & 52.4 & 34.3 & 58.8 & 64.0 \\
Ours (ADTL) & 76.3 & 66.9 & 81.2 & 81.0 & 72.4 & 62.4 & 78.5 & 76.6 & 55.4 & 35.6 & 63.0 & 67.7 \\
\hline
Human & 94.2 & 93.7 & 92.3 & 96.3 & 85.8 & 81.0 & 86.7 & 89.4 & 53.3 & 33.9 & 55.9 & 68.1 \\
\hline
\end{tabular}%
}
\end{table*}

\section{Conclusion}
This study introduces a novel embodied reference understanding framework that leverages an attention-dynamic touch line approach to enhance gesture interpretation performance. Our framework addresses the critical aspect of pointer-referent distance awareness in visual perspective-taking within the context of pointing gesture comprehension. This approach significantly reduces misinterpretations of pointing gestures across diverse interactive scenarios. We have developed a new dynamic attention source prediction mechanism based on the Virtual Touch Line concept, which adapts to varying interaction distances between the pointer and the referent object. This innovation has led to substantial improvements in gesture interpretation accuracy while simultaneously reducing model training costs. Our experimental results demonstrate significant performance enhancements in visual grounding for referring expression comprehension tasks under embodied reference conditions.

The proposed framework shows promise for application in various domains involving interpersonal communication and human-robot interaction. However, we acknowledge certain limitations in our current work, primarily the focus on pointing gestures as the sole form of nonverbal communication. Future research will expand the framework to incorporate other nonverbal cues such as gaze direction and hand gestures. We also plan to investigate multimodal integration techniques to combine various nonverbal signals for more robust embodied reference understanding. 

\section*{Acknowledgments}
This work is founded by National Natural Science Foundation of China (No.62076080, 62306083), Natural Science Foundation of ChongQing CSTB2022NSCQ-MSX0922 and the Postdoctoral Science Foundation of Heilongjiang Province of China (LBH-Z22175).

\bibliographystyle{IEEEtran}
\bibliography{ref}

\newpage
% \end{thebibliography}

\vfill

\end{document}